\documentclass[10pt,twocolumn,letterpaper]{article}

\usepackage{iccv}
\usepackage{times}
\usepackage{epsfig}
\usepackage{graphicx}
\usepackage{amsmath}
\usepackage{amssymb}
\usepackage{float} 
\usepackage{subfigure}

\usepackage[ruled,vlined]{algorithm2e}
\usepackage{listings}

\newcommand*{\thead}[1]{%
\multicolumn{1}{c}{\begin{tabular}{@{}c@{}}#1\end{tabular}}}
\usepackage{multirow}

\usepackage{tikz}
\usepackage{pgfplots}
\usetikzlibrary{patterns}

\usepackage[para]{footmisc}

\usepackage[pagebackref=true,breaklinks=true,letterpaper=true,colorlinks,bookmarks=false]{hyperref}

\iccvfinalcopy 

\def\iccvPaperID{4290} 

\ificcvfinal\fi

\begin{document}

\title{Self-Supervised Visual Representations Learning by Contrastive Mask Prediction}

\author{

Yucheng Zhao\footnotemark[1] \footnotemark[2] $^{1}$ \qquad Guangting Wang\footnotemark[1] \footnotemark[2] $^{1}$ \qquad Chong Luo$^{2}$ \qquad Wenjun Zeng$^{2}$ \qquad Zheng-Jun Zha\footnotemark[3] $^{1}$\\
University of Science and Technology of China$^{1}$ \qquad Microsoft Research Asia$^{2}$ \\
{\tt\small \{lnc, flylight\}@mail.ustc.edu.cn \quad \{cluo, wezeng\}@microsoft.com\quad zhazj@ustc.edu.cn}

}

\maketitle
\ificcvfinal\thispagestyle{empty}\fi

\renewcommand{\thefootnote}{\fnsymbol{footnote}}
\footnotetext[1]{Equal contribution.}
\footnotetext[2]{Interns at MSRA.}
\footnotetext[3]{Corresponding author.}
\renewcommand{\thefootnote}{\arabic{footnote}}
\begin{abstract}
Advanced self-supervised visual representation learning methods rely on the instance discrimination (ID) pretext task. We point out that the ID task has an implicit semantic consistency (SC) assumption, which may not hold in unconstrained datasets. In this paper, we propose a novel contrastive mask prediction (CMP) task for visual representation learning and design a mask contrast (MaskCo) framework to implement the idea. MaskCo contrasts region-level features instead of view-level features, which makes it possible to identify the positive sample without any assumptions. To solve the domain gap between masked and unmasked features, we design a dedicated mask prediction head in MaskCo. This module is shown to be the key to the success of the CMP. We evaluated MaskCo on training datasets beyond ImageNet and compare its performance with MoCo V2 \cite{DBLP:journals/corr/abs-2003-04297}. Results show that MaskCo achieves comparable performance with MoCo V2 using ImageNet training dataset, but demonstrates a stronger performance across a range of downstream tasks when COCO or Conceptual Captions are used for training. MaskCo provides a promising alternative to the ID-based methods for self-supervised learning in the wild. 
\end{abstract}

\section{Introduction}

Self-supervised learning (SSL) of visual representation has been a great success in recent years \cite{DBLP:conf/cvpr/He0WXG20,DBLP:conf/icml/ChenK0H20,DBLP:conf/nips/GrillSATRBDPGAP20,DBLP:conf/nips/CaronMMGBJ20,DBLP:conf/eccv/CaronBJD18}, facilitating a wide range of downstream tasks, including both visual tasks \cite{DBLP:conf/cvpr/HeZRS16,DBLP:journals/pami/RenHG017,DBLP:conf/iccv/HeGDG17,DBLP:journals/tomccap/LiuZCWZ19,DBLP:journals/tomccap/XieFZYLZ19}, and cross-modality tasks \cite{DBLP:conf/cvpr/ZhangSY0WHZ20,zha2019context}. Recently, several contrastive learning based SSL methods have demonstrated strong results on downstream image classification task \cite{DBLP:conf/nips/GrillSATRBDPGAP20}, closing the performance gap between them and supervised learning methods. As SSL methods do not require human annotation of training datasets, people are optimistic about the day when they surpass supervised feature learning methods when virtually unlimited data are made use of. 

\begin{figure}[t]
\centering  
\subfigure[An example image that satisfies the SC assumption.]{
\label{fig:sc_assump.1}
\includegraphics[width=0.46\linewidth]{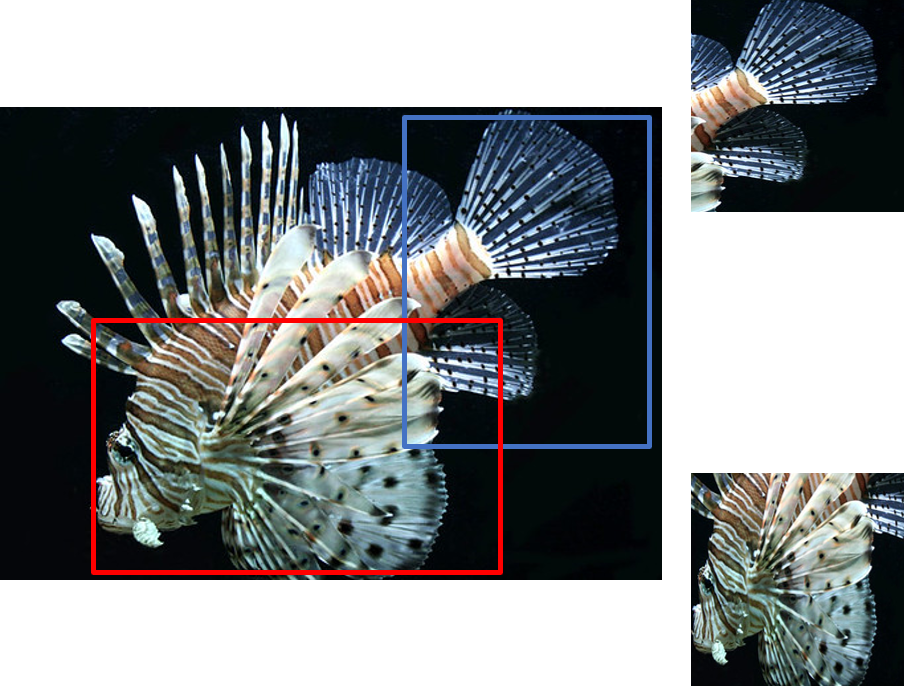}}
~
\subfigure[An example image that does not satisfy the SC assumption.]{
\label{fig:sc_assump.2}
\includegraphics[width=0.46\linewidth]{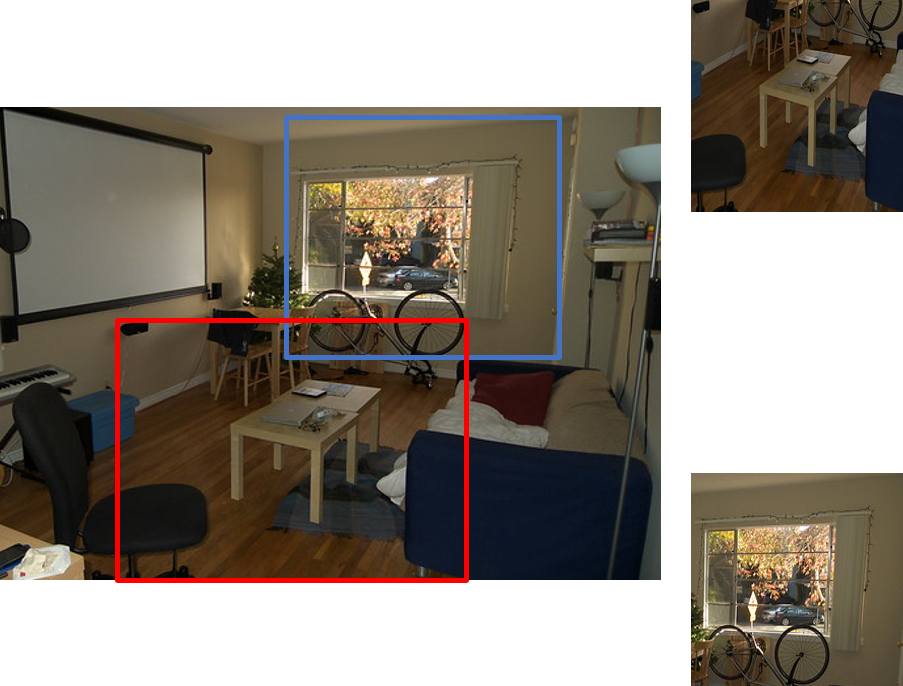}}
\caption{Illustration of semantic consistency (SC) assumption by visualizing two cropped views of an image. (a) An image from the ImageNet dataset satisfies the SC assumption, as the two cropped views sample different parts of the same fish. (b) An image from the COCO dataset does not satisfy the SC assumption, as the two cropped views contain different objects belonging to different semantic categories.}
\label{fig:sc_assump}
\end{figure}

\begin{figure}[t]
\centering  
\includegraphics[width=0.9\linewidth]{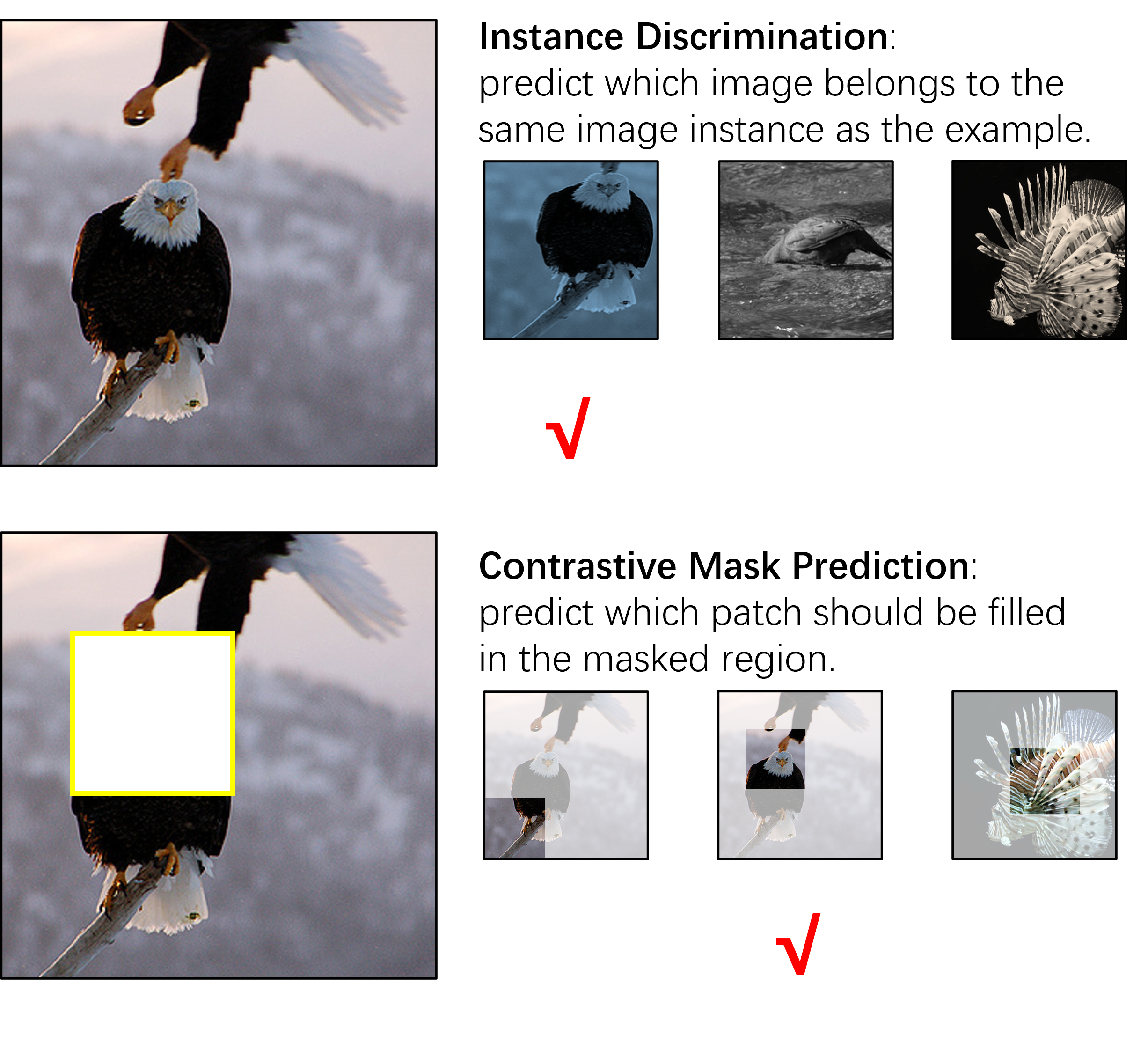}
\caption{Illustration of pretext tasks for SSL: the commonly used instance discrimination task and the proposed contrastive mask prediction task.}
\label{fig:cmp}
\end{figure}

The state-of-the-art (SOTA) SSL methods for visual feature learning are mostly based on a pretext task called instance discrimination (ID) \cite{DBLP:conf/cvpr/WuXYL18}. The idea is illustrated in Figure \ref{fig:cmp}. The intuition behind is that we can learn a good feature representation by merely asking the feature to be discriminative of individual instances. To realize this idea, SOTA SSL methods \cite{DBLP:conf/cvpr/He0WXG20,DBLP:conf/icml/ChenK0H20} optimize a contrastive loss function which enforces the model to output sufficiently close representations for different views of a single image. Since semantic information is a principal component in visual features, an implicit assumption here is that a training image should have consistent semantic meaning across views, which we call semantic consistency (SC) assumption. 

The SC assumption is almost always satisfied in the ImageNet dataset \cite{DBLP:conf/cvpr/DengDSLL009}, which exhibits an object-centric bias \cite{DBLP:conf/nips/Purushwalkam020}. An example image from the ImageNet dataset and its two cropped views are shown in Figure \ref{fig:sc_assump.1}. The two crops sample different parts of the same fish, so it is quite reasonable to let them have similar feature representations. However, if the ID-based SSL methods are to be extended to datasets beyond ImageNet, they will need to handle images like the one as shown in Figure \ref{fig:sc_assump.2}. The two cropped views of this example image contain entirely different objects belonging to different semantic categories. Is it still reasonable or helpful to enforce similar feature representations for such crops? In other words, are ID-based SSL methods readily applicable to the unconstrained datasets without SC guarantee? 


The initial studies are not encouraging. Purushwalkam \textit{et.al.} \cite{DBLP:conf/nips/Purushwalkam020} notice that the ID pretext task takes advantage of the object-centric bias in the ImageNet dataset. 
In addition, some preliminary work \cite{DBLP:conf/cvpr/He0WXG20,DBLP:conf/nips/GrillSATRBDPGAP20,DBLP:conf/cvpr/MisraM20} tries to directly extend ID-based SSL methods from ImageNet to other datasets, including Instagram-1B \cite{DBLP:conf/eccv/MahajanGRHPLBM18}, Places365 \cite{DBLP:journals/pami/ZhouLKO018}, and YFCC100M \cite{DBLP:journals/cacm/ThomeeSFENPBL16}, but they do not get satisfactory results. These facts cast a shadow over the future of ID-based SSL methods. After all, being able to make effective use of large amounts of unconstrained data is the most competitive feature of SSL methods.


We are therefore motivated to explore alternative pretext tasks which do not rely on the SC assumption. A task that once achieved great success in natural language pre-training has entered our sight. It is the mask prediction task, also known as the masked language model (MLM) in BERT \cite{DBLP:conf/naacl/DevlinCLT19}. 
In the vision domain, image inpainting is also a mask prediction task and has been used as a pretext task for visual representation learning \cite{DBLP:conf/cvpr/PathakKDDE16}. However, image inpainting is a generative method operating in the pixel space. As Grill \textit{et al.} pointed out, such method is computationally expensive, and the high level of detail required for image generation may not be necessary for representation learning \cite{DBLP:conf/nips/GrillSATRBDPGAP20}. 


In this paper, we propose to use contrastive mask prediction as a pretext task for self-supervised visual representation learning. The idea is illustrated in Figure \ref{fig:cmp} (bottom). We make the task contrastive so that the extracted features can focus on high-level semantic meanings instead of pixel-level details. In order to realize this idea, we design a novel SSL method named Mask Contrast (MaskCo). In the design of MaskCo, we answer two basic questions as to what features to contrast and how to bridge the domain gap between features of masked and unmasked regions. First, we propose to use region-level features instead of view-level features to compute contrastive loss. This choice is the key to get rid of the SC assumption because we compare the masked and unmasked versions of the exact region instead of comparing two views. Second, as a domain gap exists between the predicted features from the masked and unmasked view, we insert a mask prediction head (MPH) into the network to bridge the gap. Both quantitative and qualitative results have demonstrated that MPH is an indispensable component in MaskCo.

Evaluations of MaskCo on a range of datasets show promising results. While MaskCo achieves comparable performance with MoCo V2 \cite{DBLP:journals/corr/abs-2003-04297} when using ImageNet as the pre-training dataset, it achieves much stronger performance than MoCo on multiple downstream tasks when the unconstrained datasets COCO \cite{DBLP:conf/eccv/LinMBHPRDZ14} or Conceptual Captions \cite{DBLP:conf/acl/SoricutDSG18} are used for pre-training. These results verify that MaskCo has relaxed the SC assumption, and has high potential to be used for learning image representations in the wild.

\section{Background and Related Work} \label{sec:2}

In this section, we first provide background on self-supervised visual representation learning and the instance discrimination task. Then we discuss related work on pre-training beyond ImageNet and the context prediction task. 

\subsection{SSL and the ID task}
SSL methods try to learn generalizable and transferable representations from unlabeled data using pretext tasks that provide free supervision. With well-designed pretext tasks \cite{DBLP:conf/iccv/DoerschGE15,DBLP:conf/eccv/NorooziF16,DBLP:conf/cvpr/PathakKDDE16,DBLP:conf/eccv/ZhangIE16,DBLP:conf/eccv/CaronBJD18,DBLP:conf/iclr/GidarisSK18,DBLP:conf/cvpr/WuXYL18}, current SSL methods \cite{DBLP:conf/nips/CaronMMGBJ20,DBLP:conf/nips/GrillSATRBDPGAP20} are able to achieve comparable performance on a range of downstream tasks with their supervised counterparts. The most attractive feature of SSL is its potential to make use of unlimited data. The community are generally optimistic about its future performance when a much larger dataset than ImageNet is used for training.

Recently, the instance discrimination (ID) task \cite{DBLP:conf/cvpr/WuXYL18} has dominated SSL for its strong performance on almost all testing benchmarks \cite{DBLP:conf/cvpr/He0WXG20,DBLP:conf/icml/ChenK0H20,DBLP:conf/nips/CaronMMGBJ20,DBLP:conf/nips/GrillSATRBDPGAP20,DBLP:journals/corr/abs-2003-04297}. This task aims to discriminate each image instance by viewing an image as a distinct class of its own. The common practice to solve the ID task is to use contrastive learning. Chen \textit{et al.} \cite{DBLP:conf/icml/ChenK0H20} proposed a simple framework for contrastive learning, named SimCLR. It learns representations by contrasting images after the composition of data augmentations, showing that SOTA performance can be achieved by contrastive learning without specialized architectures or a memory bank. MoCo \cite{DBLP:conf/cvpr/He0WXG20} and its improved version MoCo V2 \cite{DBLP:journals/corr/abs-2003-04297} proposed to use momentum encoder to release the requirement of large batch-size or memory bank. More recently, BYOL \cite{DBLP:conf/nips/GrillSATRBDPGAP20} removed negative samples from contrastive learning by iteratively bootstrapping the outputs of a network to serve as targets for an enhanced representation. Meanwhile, ID task is also extended to pixel-level contrastive learning \cite{DBLP:journals/corr/abs-2011-10043,DBLP:journals/corr/abs-2011-09157,DBLP:conf/nips/PinheiroABGC20} for dense prediction tasks. These methods, however, are beyond the scope of our discussion as they generally do not perform well in image classification task. 

\subsection{Pre-training beyond ImageNet}
Although ID-based SSL methods have achieved revolutionary progress, their success is mainly confirmed on the ImageNet pre-training dataset, a labeled and curated dataset. Purushwalkam \textit{et al.} \cite{DBLP:conf/nips/Purushwalkam020} noticed that the advance of current ID-based SSL methods moderately comes from their usage of dataset bias of ImageNet. They also found that training MoCo on the less-biased COCO dataset does not get encouraging results. Xiao and Michael \cite{DBLP:conf/nips/ZhangM20} also reported degraded performance when training MoCo on COCO and Pascal datasets.

Other work \cite{DBLP:conf/nips/CaronMMGBJ20,DBLP:conf/cvpr/He0WXG20} reported preliminary results on non-ImageNet datasets. Using larger uncurated datasets of Internet images, directly applying current SSL methods has shown marginal gains, even though the dataset size is orders of magnitude larger than ImageNet. Furthermore, the current SOTA method BYOL \cite{DBLP:conf/nips/GrillSATRBDPGAP20} even shows a significant performance drop when replacing the pre-training dataset from ImageNet to Place365. While it is still not clear whether the performance drop is due to the domain gap between pre-training and downstream datasets, it is clear that the ID task makes a strong SC assumption on the training images, which can hardly hold in unconstrained datasets.

\subsection{Context prediction tasks}
In this paper, we propose a novel pretext task named Contrastive Mask Prediction. It is inspired by the MLM in BERT \cite{DBLP:conf/naacl/DevlinCLT19} and has a strong correlation with some context-based prediction tasks \cite{DBLP:conf/iccv/DoerschGE15,DBLP:conf/eccv/NorooziF16,DBLP:conf/cvpr/PathakKDDE16} in visual representation learning.
BERT \cite{DBLP:conf/naacl/DevlinCLT19} is a milestone pre-training method in natural language processing. In the core of BERT is the mask prediction task, also known as the mask language model (MLM). From the input sentence, some tokens are randomly masked out, and the training objective is to predict the vocabulary ID of the masked word based on its context. MLM cannot be directly extended to handle images as images are continuous signals without a finite dictionary. 

In the computer vision domain, RelativePosition \cite{DBLP:conf/iccv/DoerschGE15} is a context-based pretext task whose goal is to predict the relative position between two image patches. The authors hypothesized that doing well on this task requires the  understanding of scenes and objects. Following this idea, Noroozi \textit{et al.} \cite{DBLP:conf/eccv/NorooziF16} proposed a self-supervised learning method based on solving jigsaw puzzles, reducing the ambiguity of the pretext task, and achieving better performance on multiple downstream tasks. However, these tasks have relied on heuristics which could limit the generality of the learned representations \cite{DBLP:conf/icml/ChenK0H20}.

Image inpainting \cite{DBLP:conf/cvpr/PathakKDDE16} is also used as a pretext task in SSL by hypothesizing that deep semantic understanding is essential to recover the missing content. As we have pointed out earlier, this work falls into the category of generative methods. In this work, we embrace the contrastive methods and the reasons will be detailed in the next section.

\section{Learning by Contrastive Mask Prediction}

In this section, we first introduce the CMP task, which relaxes the SC assumption in SSL and enables the representation learning in the wild. Then we describe our approach called mask contrast (MaskCo), which concretely implements the CMP task. Last, the implementation details are presented.

\subsection{The Contrastive Mask Prediction Task} \label{sec:4.1}

The mask prediction task explores the data structure through the context correlation. Given a natural signal, a small region is masked out before it is processed by a neural network model. The training objective is to obtain a feature representation of the input signal, from which the exact signal of the masked region or its feature can be predicted. The mask prediction task involves a fundamental assumption, that is, correlation exists between content and its context, which we believe is generally valid for natural signals, including language and images. However, the prediction process is easy to implement for tokenized data like text but not so straightforward for continuous data like images. To adapt the mask prediction on the image, we modify the objective from predicting a permanent category ID to distinguishing the true candidates from distractions and implement it by the contrastive loss. We thus name the modified task as contrastive mask prediction (CMP).

\begin{figure*}[th]
\begin{center}
\includegraphics[width=0.9\linewidth]{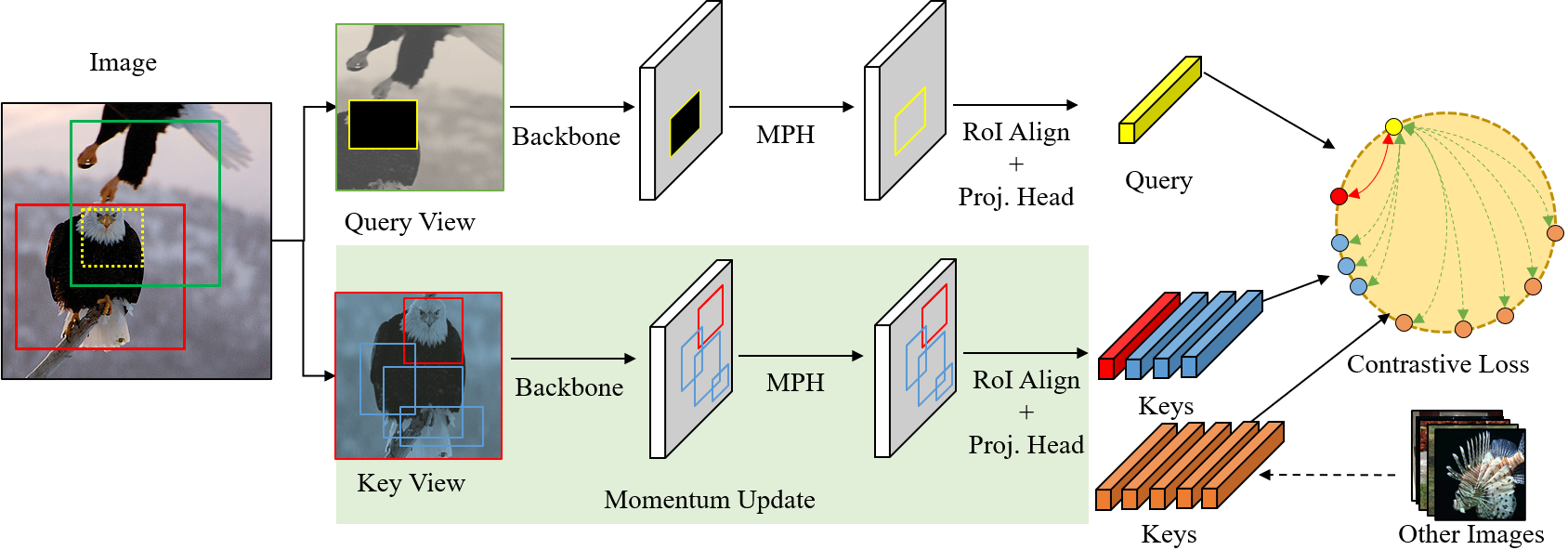}
\end{center}
   \caption{Illustration of \textbf{Mask Contrast (MaskCo):} Given a training image, we first acquire two views by data augmentations. A selected region is masked out in the query view while the key view stays unaltered. We obtain the query and key features by a forward pass through the backbone network, the mask prediction head (MPH), and the RoI align layer with a projection head. We perform region-level contrastive learning, in which the query feature for the masked region is pulled closer to the positive key feature, and pushed away from the negative key features. Most negative keys are from other images while we could also include a few negative keys from other regions of the same image.}
\label{fig:framework}
\end{figure*}

An illustration of CMP is shown in Figure \ref{fig:cmp} (bottom). The image contains a main object which is a bird, and the bird's top body is masked out. Looking at the masked image, people can reasonably guess that the masked region should include a bird's head. We come up with this guess by looking at the surrounding context where a twig and the bird's legs appear. This ability comes from understanding the semantic content, and it is indeed what we want a pre-trained representation to possess. However, it will be hard for a person to draw the exact pixel of the missing part since the details of a bird could vary. Likewise for a network, filling in the missing part in the pixel-level will be computationally expensive and unnecessary. This is why we choose to use a contrastive version of mask prediction instead of a generative version as used in previous work \cite{DBLP:conf/cvpr/PathakKDDE16}.

\subsection{The Mask Contrast Approach} \label{sec:4.2}
This subsection describes the Mask Contrast (MaskCo) approach we developed to realize the CMP task. Figure \ref{fig:framework} provides an overview of the approach. Similar to MoCo, we adopt a Siamese network structure in pre-training, the online network processes the query view and the target network processes the key view. The parameters in the online network are updated through back-propagation, while those in the target network are updated with momentum. 

The first prominent feature of MaskCo is that we perform region-level contrast instead of view-level contrast. The region refers to the masked region in the image and all of the positive and negative samples. While this idea is quite straightforward to the mask prediction task, we would like to emphasize that it is the key to release the SC assumption used in ID tasks. This is because, for the predicted feature of the masked region, the representation of the same region drawn from an unmasked view is an indisputable positive sample that does not depend on any hypothesis. 


Given an image instance, we first create query view and key view through random cropping and other data augmentation techniques. The constraint here is that the two views should have a sufficiently large overlap. Then, a masked region is identified within the overlap of the two views. In the query view, the selected region is masked to zero. In the key view, the selected region is the only positive key box. We additionally pick several negative key boxes, which have a small overlap with the positive key box. More negative key boxes are sampled from other images. The query view and the key view are passed through the backbone network, and the region-level feature is extracted by RoI align operation after the features pass through an additional mask prediction head (MPH). The final feature to compute contrastive loss is the RoI feature passing a non-linear projection head \cite{DBLP:journals/corr/abs-2003-04297, DBLP:conf/icml/ChenK0H20}.



The second prominent feature of MaskCo is the design of MPH. We rely on MPH to bridge the domain gap between features of masked and unmasked regions. In most of the downstream tasks, the backbone network is used to acquire representations for unmasked images. Therefore, it does not need to have the filling-in-the-blank capability. However, in the pre-training stage, we contrast the feature of the masked region with all other features of unmasked regions. The network will naturally learn how to do mask prediction. There is a domain gap between the generally useful visual features (of unmasked regions) and the task-specific features (of masked regions). We therefore design a specialized module for mask prediction and insert it between the backbone network and the RoI align operation. The MPH is simply implemented by a few convolution layers. Later, our visualization and quantitative evaluation in Section \ref{sec:5.4} will confirm its effectiveness. 

We use contrastive loss \cite{DBLP:conf/cvpr/He0WXG20,DBLP:journals/corr/abs-1807-03748} as our training objective, which can be thought of as a \textit{dictionary look-up} task. Consider an encoded feature $q$ as the query and a set of encoded features $\{k_i|i=0,...,K\}$ as keys of a dictionary. Assume there is a single key (denoted as $k^+$) that matches the query $q$. The contrastive loss is a function whose value is low when $q$ is similar to the positive key and dissimilar to all other keys. With similarity measured by  L2-normalized dot product, the loss function can be formulated in the following:
\begin{equation} \label{eq:1}
    L = -\log{\frac{\mathrm{exp}(q\cdot k^+ / \tau)}{\sum_{i=0}^{K}{\mathrm{exp}(q\cdot k_i / \tau)}}}
\end{equation}
where $\tau$ is a temperature hyper-parameter \cite{DBLP:conf/icml/ChenK0H20,DBLP:conf/cvpr/He0WXG20}. This form is the exact one appeared in \cite{DBLP:conf/cvpr/He0WXG20}, while it can also take other forms \cite{DBLP:conf/icml/ChenK0H20,DBLP:journals/corr/abs-1807-03748}.

In MaskCo, the query $q$ is the region-level feature of the masked box, the positive key $k^+$ is the feature of the corresponding region in the key view. The negative keys ${k_i|i=0,...K,k_i\neq k^+}$ are the feature vectors of negative key boxes drawn from other images or the same image. It is worth noting that contrasting region-level features allows us to include negative samples within the same image. This is not possible in ID-based methods because they use view-level features. We conjecture that negative keys from the same image can be viewed as hard negatives which might enhance the localization capability of the trained features. We therefore present two variants of MaskCo. The default version only uses inter-image negatives and the MaskCo(+in) uses both inter-image and intra-image negatives. We show that MaskCo(+in) has a small advantage in some detection benchmarks.

\subsection{Implementation Details} \label{sec:4.3}

Next we provide implementation details of MaskCo.

\textbf{Data augmentation details}: The global image augmentation on the key view and the query view is the same as MoCO v2 \cite{DBLP:journals/corr/abs-2003-04297}, except the random crop is constrained as described in Section \ref{sec:4.2}. Given an image, a random region is cropped with at least 20\% of the image and resized to 224 × 224 with a random horizontal flip, followed by a random color jittering related to brightness, contrast, saturation, hue, and grayscale. Gaussian Blur is also used for augmentation. After global augmentation, we randomly generate 17 bounding boxes in the key view with the size from 32x32 to 128x128. The first one is viewed as the positive key box, and others are viewed as the negative key boxes. The projection of the positive key box in the query view is viewed as the masked box. We enforce the negative key boxes in the generation process to have an IoU lower than 0.2 with the positive key box.

\textbf{Model details:} We use the OpenSelfSup\footnote{https://github.com/open-mmlab/OpenSelfSup} codebase to implement MaskCo. The backbone model is standard ResNet-50 \cite{DBLP:conf/cvpr/HeZRS16}. The mask prediction head is three 3-layer blocks of residual layers (1x1 conv-3x3 conv-1x1 conv), which is the same as conv5 layer in the ResNet-50. We use the same projection head as MoCo v2 \cite{DBLP:journals/corr/abs-2003-04297}.

\textbf{Training details:} The training epoch is set to 100 in most experiments on ImageNet dataset, and set to the same iterations on other datasets. For comparison with other methods on ImageNet (in Table \ref{tab:2}), we set the pre-training epoch to 200. Most training hyper-parameters are the same as MoCo V2 \cite{DBLP:journals/corr/abs-2003-04297}. Specifically, we use the SGD optimizer and set the base learning rate as 0.03, momentum as 0.9, weight decay as 0.001, and total batch size as 256 distributed across 8 V100 GPU. The cosine learning rate schedule is also used. The training details for the downstream tasks can be found in Appendix.

\section{Experiments}

\subsection{Experiment Setup}

We use four datasets in our experiment. The first one is the COCO dataset \cite{DBLP:conf/eccv/LinMBHPRDZ14}, which has 328k images and is commonly used for complicated scene understanding tasks, including object detection and semantic segmentation. The second one is the Conceptual Captions (CC) dataset \cite{DBLP:conf/acl/SoricutDSG18}, which is designed for image captioning. CC dataset contains 3.3 million images acquired from the Internet. Both COCO and CC do not exhibit the object-centric bias as ImageNet. For completeness and comparison with other SSL methods, we also include ImageNet (IN) dataset \cite{DBLP:conf/cvpr/DengDSLL009}, which has 1.3 million images belonging to 1000 classes. IN is well-balanced in its class distribution and each image generally contains iconic views of objects. The last dataset is ImageNet 10\% (IN-10\%), which is constructed by sampling 1/10 of the images per class on the original ImageNet.

We use four downstream tasks for evaluation. 

\textbf{ImageNet Linear Classification:} We follow the linear evaluation protocol \cite{DBLP:conf/iccv/GoyalM0M19}. When we train on the ImageNet dataset, the pre-trained model is fixed, and only the additional linear classification layer is fine-tuned. We report the top-1 accuracy of pre-trained features from conv4 and conv5, since most pretext tasks achieve the best performance on these two layers. The full results can be found in the Appendix.

\textbf{Pascal VOC Object Detection:} We follow the setting introduced in MoCo \cite{DBLP:conf/cvpr/He0WXG20}. A Faster R-CNN detector \cite{DBLP:journals/pami/RenHG017} with the ResNet50-C4 backbone is implemented in Detectron2 \cite{wu2019detectron2}. We fine-tune all layers end-to-end and synchronize all batch normalization layers. The fine-tuning is performed on \textit{trainval07+12} set (16.5k images) and the evaluation is on \textit{test2007} set. We report evaluation metrics AP, AP50, and AP75.

\textbf{COCO Object Detection and Instance Segmentation:} We follow the setting introduced in MoCo \cite{DBLP:conf/cvpr/He0WXG20}, where a Mask RCNN detector with the ResNet50-C4 backbone is used, implemented in Detectron2 \cite{wu2019detectron2}. In the optimization, we fine-tune all layers end-to-end and synchronize all batch normalization layers. We use the 2x schedule in Detectron2. The fine-tuning is performed on \textit{train2017} set (118k images) and evaluation is on \textit{val2017} set. In the evaluation, we report AP, AP50, and AP75 for detection as well as AP, AP50, and AP75 for segmentation.


\subsection{Dataset Analysis}
Previous work \cite{DBLP:conf/nips/Purushwalkam020} has suggested that ID-based SSL methods take advantage of the object-centric bias of the ImageNet dataset. This is highly related to the semantic consistency assumption we pointed out. In order to better understand this problem, 
we define an objective metric, called the \textit{ResNet distance}, to reflect the semantic consistency in an image. The major tool we use is the PyTorch official ResNet-50 model. Specifically, the model is pre-trained with the image classification task on the ImageNet dataset in a supervised way. 

Given an image, we generate two views $x_1,x_2$ using exactly the same random crop operation as in MoCo, and compute the ResNet distance as follows:
\begin{equation} \label{eq:2}
    d(x_1,x_2) = ||\hat{y}^1 - \hat{y}^2||^2,
\end{equation}
where $\hat{y}_1$ and $\hat{y}_2$ are conv5 features, with unit-normalization in the channel dimension, of $x_1$ and $x_2$ generated by the aforementioned ResNet-50 model.

We calculate the average ResNet distance for various pre-training datasets, including ImageNet, ImageNet 10\%,  COCO and CC. We can find from Table \ref{tab:0} that the ResNet distances of COCO and CC are significantly larger than those of ImageNet datasets, suggesting that the SC assumption does not hold well on non-ImageNet datasets.

\begin{figure}
    \centering
    \begin{tikzpicture}
    \begin{axis}[%
    width=8cm,
    height=4cm,
    compat=1.3,
        symbolic x coords={1,2,3},
    	enlargelimits=0.15,
    	legend style={legend columns=-1
    	},
    	legend pos=north west,
    	ymax=62,
    	ybar=9pt,
    	xtick=data,
    	xticklabels={NPID, SimCLR, MoCo v2},
    	bar width=9pt,
    	area legend,
    	y label style={anchor=south},
    	ylabel={ImageNet Top-1 Acc (best)},
    	label style={font=\scriptsize}
    	]

        \addplot[ybar, pattern=dots] coordinates {(1,44.9) (2,55.6) (3,58.8)};
        
        \addplot[ybar, pattern=horizontal lines] coordinates {(1,43.0) (2,51.2) (3,55.6)};
        \legend{IN-10\%, COCO}
    \end{axis}
\end{tikzpicture}
    \caption{Pre-training three ID-based SSL methods on IN-10\% and COCO. The ImageNet linear classification top-1 accuracy from best layers are reported. Significant performance drop is observed when training on COCO.}
    \label{fig:0}
\end{figure}
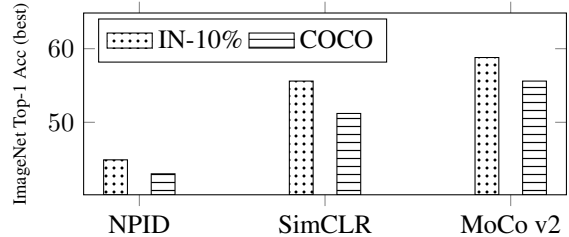

Furthermore, we show that the performance of ID-based SSL methods drops significantly on pre-training datasets with a larger ResNet distance. In order to eliminate the dataset size factor, we carry out the comparison using COCO dataset and ImageNet 10\% dataset, whose sizes are almost identical. Three SOTA ID-based SSL methods, including NPID, SimCLR, and MoCo v2, are evaluated. The results are shown in Figure \ref{fig:0}. We can find that all ID-based methods have significant performance drop when transferring to non-ImageNet datasets.

\begin{table}
\begin{center}
\begin{tabular}{ccccc}
\hline
\thead{Dataset}           & IN & IN-10\% & COCO   & CC \\
\hline
\thead{ResNet Dist.} &  0.1546  & 0.1547                  & 0.1814 & 0.1782 \\
\hline
\end{tabular}
\end{center}
\caption{ResNet distance for multiple pre-training datasets. The distances of non-ImageNet datasets are significantly larger than ones in ImageNet dataset.}
\label{tab:0}
\end{table}

\subsection{Main Results}

We provide our experimental results for object detection, image classification, and instance segmentation tasks, compare MaskCo's performance with state-of-the-art SSL methods. All compared methods use the same training settings and are implemented by the OpenSelfSup.

The main advantage of MaskCo over ID-based methods is that it does not rely on the SC assumption. In order to verify this, we train MaskCo and MoCo v2 using different pre-training datasets and compare their performance on VOC detection and ImageNet linear classification. The results are shown in Table \ref{tab:1}. When pre-trained on ImageNet, MaskCo achieves slightly lower performance on three benchmark tasks. However, for the other two pre-training datasets, MaskCo achieves consistently higher performance than MoCo v2. For COCO dataset, which has the largest ResNet distance among the three datasets, MaskCo outperforms MoCo v2 with 0.6 AP on Pascal detection, and 1.1\% on ImageNet linear classification (Comparison based on the best result from conv4 and conv5). For Conceptual Captions, which is another unconstrained dataset, MaskCo achieves a similar performance gain over MoCo on all three tasks. 

We noticed that although MaskCo shows relative advantage over MoCo v2 on COCO and CC datasets, there is a performance gap between the model trained on COCO/CC and the model trained on ImageNet. One possible reason is that we keep all the hyper-parameters tuned on ImageNet for a fair comparison among methods, but this is definitely unfair when it comes to the performance comparison among pre-training datasets. In the future, we plan to explore automatic hyper-parameter adjustment policy for unconstrained dataset, but it is beyond the scope of this work.

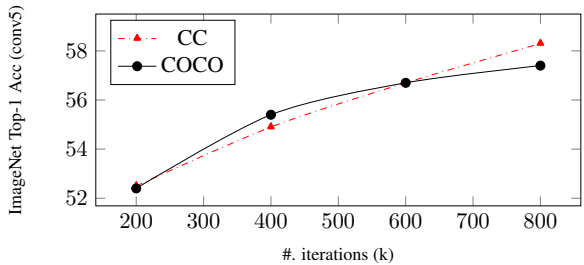
\begin{figure}[t]
    \centering
    \begin{tikzpicture}[scale=0.85]
    \begin{axis}[
    xlabel={\#. iterations (k)},
    ylabel={ImageNet Top-1 Acc (conv5)},
	ymax=59.5,
    height=0.55\linewidth,
    width=1.1\columnwidth,
    legend pos=north west,
    label style={font=\footnotesize}
    ]
    \addplot[dash dot, smooth,mark=triangle*, color=red] plot coordinates {
        (200, 52.5)
        (400, 54.9)
        (600, 56.7)
        (800, 58.3)
 
    };
    \addlegendentry{CC }
    
    \addplot[solid, smooth,mark=*, color=black] plot coordinates {
        (200, 52.4)
        (400, 55.4)
        (600, 56.7)
        (800, 57.4)
 
    };
    \addlegendentry{COCO }
    \end{axis}
    \end{tikzpicture}
    \caption{Comparison of pre-trained MaskCo with ImageNet linear classification top-1 accuracy across different pre-training iterations on COCO and CC datasets. Pre-training on COCO gets saturated performance after 600k iterations, while the model pre-trained on CC can continuously improve.}
    \label{fig:5}
    \end{figure}

Besides, although CC is a much larger dataset than COCO, the results of MaskCo trained on CC do not show much advantage over the results trained on COCO in Table \ref{tab:1}. This is reasonable because we are showing the results achieved at the same iteration steps. To make the point clear, we conduct side-by-side experiments on these two datasets and show the results at multiple training iterations from 200k to 800k (the default setting reported in Table \ref{tab:1} uses 500k iterations). As shown in Figure \ref{fig:5}, training on COCO starts to have a saturated performance after 600k iterations, but training on the larger CC dataset does not show any sign of saturation. The performance still grows almost linearly at 800k iterations. It matches the conventional wisdom that training on larger datasets needs more iteration steps to converge. 

Last, we present a comparison between MaskCo and most SOTA SSL methods in Table \ref{tab:2}. The pre-training dataset is ImageNet, so it is not a setting to show the advantage of MaskCo. But even in such a setting, MaskCo manages to outperform most other SSL methods on the detection and segmentation tasks. Only on the ImageNet classification task, MaskCo is slightly inferior to MoCo v2 and BYOL. Another variant of MaskCo, which uses the intra-image negative samples, achieves an AP of 57.2 on Pascal VOC, which is slightly better than MaskCo. But on the image classification task, it incurs some losses.  

\begin{table}
\begin{center}
\setlength\tabcolsep{4pt}
\begin{tabular}{c|c|ccc|cc}
\hline
\multirow{2}{*}{Dataset} & \multirow{2}{*}{Method} & \multicolumn{3}{c|}{VOC} & \multicolumn{2}{c}{ImageNet}  \\
     &                              & $AP$ & $AP_{50}$ & $AP_{75}$ & conv4 & conv5 \\
\hline
\multirow{2}{*}{IN}       & MoCo v2 & \textbf{56.4} & \textbf{82.0} & 62.8 & \textbf{60.0} & \textbf{64.4}\\
                          & MaskCo  & 56.0 & 81.3 & \textbf{63.1} & 57.6 & 63.1\\

\hline
\multirow{2}{*}{COCO}     & MoCo v2 & 55.0 & 80.7 & 60.9 & 55.6 & 51.4\\
                          & MaskCo  & \textbf{55.6} & \textbf{81.0} & \textbf{61.5} & \textbf{55.7} & \textbf{56.7}\\
\hline
\multirow{2}{*}{CC}       & MoCo v2 & 55.1 & 80.8 & 60.6 & \textbf{56.0} & 56.5\\
                          & MaskCo  & \textbf{55.7} & \textbf{81.0} & \textbf{61.5} & 55.4 & \textbf{57.1}\\
\hline

\end{tabular}
\end{center}
\caption{Comparison of MoCo v2 and MaskCo on multiple pre-training datasets. Evaluations are conducted on Pascal VOC object detection and ImageNet linear classification. All evaluated models use the same training steps, which is about 500k (equal to 100 epoch on ImageNet).}
\label{tab:1}
\end{table}

\begin{table*}
\begin{center}
\begin{tabular}{l|ccc|ccc|ccc|cc}
\hline
\multirow{2}{*}{Method} & \multicolumn{3}{c|}{Pascal VOC} & \multicolumn{3}{c|}{MS COCO} & \multicolumn{3}{c|}{MS COCO} & \multicolumn{2}{c}{ImageNet}\\

& $AP$ & $AP_{50}$ & $AP_{75}$ & $AP^{bb}$ & $AP^{bb}_{50}$ & $AP^{bb}_{75}$ & $AP^{seg}$ & $AP^{seg}_{50}$ & $AP^{seg}_{75}$ & conv4 & conv5\\
\hline
Rand Init   & 32.8  & 59.0  & 31.6  & 35.6  & 54.6  & 38.2  & 31.4  & 51.5  & 33.5  & 9.1 & 6.5\\
Supervised  & 54.2  & 81.6  & 59.8  & 40.0  & 59.9  & 43.1  & 34.7  & 56.5  & 36.9  & 67.6 & 76.2\\
\hline
Relative-Pos \cite{DBLP:conf/iccv/DoerschGE15}& 55.1  & 80.4  & 61.2  & 40.0  & 59.6  & 43.5  & 35.0  & 56.5  & 37.3 & 49.3 & 40.2\\
Rotation-Pred \cite{DBLP:conf/iclr/GidarisSK18}&55.5  & 80.9  & 61.4  & 40.0  & 59.3	& 43.6	& 34.9  & 56.0	& 37.4 & 55.0 & 49.1\\
\hline
NPID \cite{DBLP:conf/cvpr/WuXYL18}        & 54.1  & 80.0  & 59.5  & 39.4  & 59.0  & 42.8  & 34.5  & 55.9	& 36.6 & 57.7 & 51.7\\
MoCo v2 \cite{DBLP:journals/corr/abs-2003-04297}     & \textbf{57.0}  & \textbf{82.2}  & \underline{63.4}  & \textbf{41.0} & \textbf{60.6} & \textbf{44.5} & \textbf{35.6} & \textbf{57.2} & \textbf{38.0} & \underline{61.6} & \underline{66.7}\\
SimCLR \cite{DBLP:conf/icml/ChenK0H20}      & 51.5  & 79.4  & 55.6  & 39.6 & 59.1 & 42.9 & 34.6 & 55.9 & \underline{37.1} & 54.4 & 61.6\\
BYOL \cite{DBLP:conf/nips/GrillSATRBDPGAP20}        & 49.0  & 49.6  & 52.8  & 40.2 & \textbf{60.6} & 43.3 & 34.9	& 57.0 & 36.7 & \textbf{62.8} & \textbf{71.6}\\
\hline
MaskCo        & \underline{56.7}  & \underline{82.1}  & \textbf{63.9}  & \underline{40.8} & \underline{60.5} & \underline{44.2} & \underline{35.5} & \underline{57.1} & \textbf{38.0} & 59.6 & 65.1\\
\hline
\end{tabular}
\end{center}
\caption{Comparison of MaskCo to previous supervised/self-supervised pre-training methods on ImageNet pre-training datasets. Evaluations are conducted on Pascal VOC object detection, COCO object detection and instance segmentation, and ImageNet linear classification. All models are pre-trained with 200 epochs and batch size of 256 using an open-source implementation, OpenSelfSup. }
\label{tab:2}
\end{table*}

\subsection{Ablation Study} \label{sec:5.4}

We conduct ablation studies using the Pascal VOC object detection and ImageNet linear classification tasks. The pre-training dataset is ImageNet.

\textbf{Effect of Mask Prediction Head:} We evaluate the effectiveness of the mask prediction head introduced in Section \ref{sec:4.2}. MPH is implemented by three 3-layer blocks of residual layers in the default setting, and here, we vary the number of 3-layer blocks to inspect this design. In Table 4, the first row reports the results when the block number is 0, or no MPH is used. Compared with other rows that use MPH, we can notice a significant performance drop in the ImageNet classification task. This confirms our conjecture that a domain gap exists between the image classification task and the CMP pretext task. Another observation is the object detection APs on Pascal VOC are not influenced by MPH. We believe this is caused by the architecture of the detector we used, whose backbone ends with conv4. Other results with $\{1,2,3,4\}$ 3-layer blocks show a trend that a sufficiently large network is necessary for filling the domain gap, and three blocks are enough in our scenario.

We also visualize features before and after MPH to demonstrate the impact of MPH. We feed example image pairs into a pre-trained MaskCo model and visualize the feature maps of conv5 and MPH. In an image pair, the first is the original image, and the second is a masked image where the 64x64 region in the center is masked by zero. We find the features corresponding to the masked region in conv5 exist noticeable low response while they are recovered to a certain extent after MPH.

\begin{table}
\begin{center}
\begin{tabular}{c|ccc|cc}
\hline
\#. 3-layer & \multicolumn{3}{c|}{Pascal VOC} & \multicolumn{2}{c}{ImageNet} \\
blocks& $AP$ & $AP_{50}$ & $AP_{75}$ & conv4 & conv5 \\
\hline
 0 & 56.1 & 81.7 & 62.2 & 57.2 & 41.9\\
 \hline
 1 & 56.1 & 81.8 & 62.4 & 58.1 & 59.0\\
 2 & 56.4 & 81.8 & 62.8 & 57.9 & 62.1\\
 3 & 56.0 & 81.3 & 63.1 & 57.6 & 63.1\\
 4 & 56.0 & 81.6 & 62.0 & 57.3 & 63.0\\
 
\hline
\end{tabular}
\end{center}
\caption{Ablation studies on mask prediction head. The first row shows the results of removing MPH and other rows show the results of using different number of 3-layer blocks.}
\label{tab:5}
\end{table}

\begin{figure}[t]
\centering  
\includegraphics[width=0.9\linewidth]{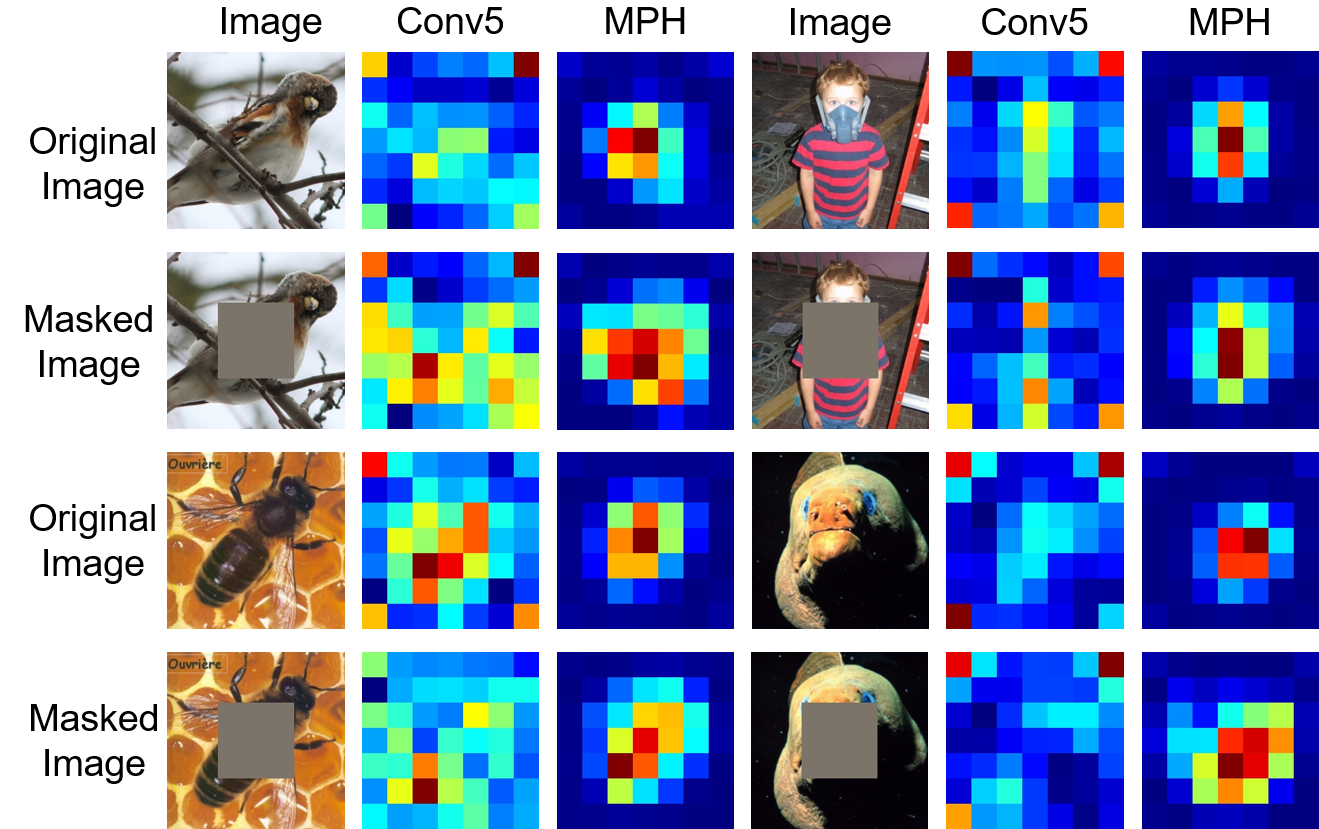}
\caption{Visualization of MPH. Given an image and its masked counterpart, we visualize their feature maps of conv5 and MPH. The features in the masked region is noticeable low in conv5 while being recovered in MPH. More visualization results can be found in Appendix.}
\label{fig:6}
\end{figure}

\textbf{Effect of Mask:} We verify that the masking operation defined in CMP is essential to learn features that transfer well. Without the mask, the pretext task will degenerate to predict a region-level feature from the exact same region with non-geometric data augmentations. This is somewhat equal to the instance discrimination task removing random cropping. Intuitively, it is an effortless task and may not drive the model to learn semantically meaningful representations.  In our practice, removing the mask indeed weakens the pretext task, makes the loss decrease from 2.51 to 1.83, and causes performance drop in the downstream tasks, as shown in Table \ref{tab:4}.



\begin{table}
\begin{center}
\begin{tabular}{c|ccc|cc}
\hline
\multirow{2}{*}{Mask}   & \multicolumn{3}{c|}{Pascal VOC} & \multicolumn{2}{c}{ImageNet}  \\
                        & $AP$ & $AP_{50}$ & $AP_{75}$    & conv4 & conv5 \\
\hline
\checkmark              & 56.1 & 81.7 & 62.2 & 57.2 & 41.9\\
                        & 53.6 & 79.4 & 59.3 & 51.9 & 52.3\\
\hline
\end{tabular}
\end{center}
\caption{Ablation studies on the mask strategy. MPH is not used in these experiments.}
\label{tab:4}
\end{table}

\textbf{Negative Sampling Strategy:} We design a novel negative sampling strategy in Section \ref{sec:4.2}, where not only the inter-image negatives are taken into account but intra-image negatives may also be included. The inner-image negatives are designed to enhance pre-trained representations' localization capability since they are harder to discriminate. We have mentioned that adding intra-image negatives improves the object detection AP, but degrades the image classification accuracy. We believe this is caused by the inconsistency between the two types of negatives, confusing the network about how much difference with the positive is sufficient to be classified into negative. We would like to explore a more elegant fusion strategy for the two types of negatives in future work. Moreover, we show that only using intra-image negative is not enough to learn good representations, which is caused by insufficient diversity of intra-image negatives.


\begin{table}
\begin{center}
\begin{tabular}{c|c|ccc|cc}
\hline
\multirow{2}{*}{inter} & \multirow{2}{*}{intra} & \multicolumn{3}{c|}{Pascal VOC} & \multicolumn{2}{c}{ImageNet}  \\
& & $AP$ & $AP_{50}$ & $AP_{75}$ & conv4 & conv5 \\
\hline
\checkmark &            & 56.0 & 81.3 & 63.1 & 57.6 & 63.1\\
           & \checkmark & 55.6 & 80.7 & 61.5 & 50.8 & 48.6\\
\checkmark & \checkmark & 56.4 & 81.6 & 63.1 & 57.5 & 61.3\\
\hline
\end{tabular}
\end{center}
\caption{Ablation studies on the negative sampling strategy. We try the inter-image negatives and intra-image negatives as well as their combination. Results reported at 100 epoch.}
\label{tab:3}
\end{table}

\section{Conclusion}

In this work, we propose a novel SSL pretext task, named contrastive mask prediction (CMP), and the framework to implement CMP, named mask contrast (MaskCo). CMP is motivated by the analysis of the limitation in current dominant pretext task, called instance discrimination (ID). We point out that ID is based on a semantic consistency assumption of curated data, leading to deteriorated performance when trained on unconstrained datasets. Our approach is based on a more fundamental assumption that contexts are correlated. It achieves consistent performance gain on multiple pre-training datasets over an ID-based method MoCo v2. As a generic pretext task, CMP has shown great promise towards training on unconstrained datasets, leading a new path to the ultimate goal of self-supervised learning: training with unlimited data in the wild.

\appendix
\section{Additional Implementation Details}

\subsection{Pseudo Code for MaskCo Training Loop}

We provide the pseudo code for MaskCo training loop in Algorithm \ref{alg:train}.

\begin{algorithm*}[h]
\caption{Pseudocode for MaskCo training loop.}
\label{alg:train}
\definecolor{codeblue}{rgb}{0.25,0.5,0.5}
\lstset{
  backgroundcolor=\color{white},
  basicstyle=\fontsize{7.2pt}{7.2pt}\ttfamily\selectfont,
  columns=fullflexible,
  breaklines=true,
  captionpos=b,
  commentstyle=\fontsize{7.2pt}{7.2pt}\color{codeblue},
  keywordstyle=\fontsize{7.2pt}{7.2pt},
}
\begin{lstlisting}[language=python]
# net_q: encoder for query image, including the backbone and MPH
# net_k: momentum encoder for key image
# head_q: projection head for query image
# head_k: projection head for key image
# m: momentum
# t: temperature
# x: input image

# generate two views, masked box, and key boxes
x_q, x_k, masked_box, key_boxes = transform(x) 

q = head_q(roi_align(net_q(x_q), masked_box) # queries: Nx1xC
k = head_k(roi_align(net_k(x_k), key_boxes) # keys: NxKxC
k = k.detach() # no gradient to keys

pos_k = k[:, 0:1]
neg_k_inter = sample_inter(k) # inter-image negatives: NxK1xC
neg_k_intra = sample_intra(k) # intra-image negatives: NxK2xC
neg_k = cat([neg_k_inter, neg_k_intra], dim=1)


l_pos = bmm(q, pos_k.transpose(1, 2)) # positive logits: Nx1
l_neg = bmm(q, neg_k.transpose(1, 2)) # negative logits: Nx(K1+K2)
logits = cat([l_pos, l_neg], dim=1)

# MoCo contrastive loss (Positive labels at index 0). 
labels = zeros(N) 
loss = CrossEntropyLoss(logits/t, labels)

# SGD update: query network
loss.backward()
update(net_q.params)
update(head_q.params)

# momentum update: key network
net_k.params = m*net_k.params+(1-m)*net_q.params
head_k.params = m*head_k.params+(1-m)*head_q.params
\end{lstlisting}
\end{algorithm*}

\subsection{Additional Training Details for Pre-training}

\textbf{Negative Sampling Strategy:} In the pre-training stage, our default model uses $M\times N=16\times31=496$ negative samples from the same GPU, where $M=16$ is the number of negative boxes per image, and $N=31$ is the number of images per GPU minus one (excluding the same image). The MaskCo(+in) model uses additional intra-image negatives, which is additional $16$ negative samples from the same image so that the number of total negative samples become $512$.

\textbf{Pre-training on Multiple Datasets:} In all pre-training datasets, including ImageNet, CC, and COCO, we use exactly the same training hyper-parameters that is tuned on the ImageNet.

\subsection{Training Details for Downstream Tasks}

\textbf{ImageNet Linear Classification:} We adopt the ImageNet Linear Classification protocol used in \cite{DBLP:conf/iccv/GoyalM0M19}. The features from different residual layers, from conv1, conv2, until conv5, are extracted, and additional pooling and linear layers are added on top of the extracted features. Only the additional linear layer is trainable. We call this evaluation protocol as \textit{multi-layer linear evaluation}. There is another linear evaluation protocol used in MoCo \cite{DBLP:conf/cvpr/He0WXG20} in which only the global average pooling features were used to train the linear classifier. We call this evaluation protocol as \textit{last-layer linear evaluation}. We observe that the last-layer protocol is biased towards the ID-based methods, and SSL methods based on other pretext tasks usually do not produce the best result at the final layer. Since we intend to develop a new pretext task, we choose the \textbf{multi-layer} protocol to evaluate the capabilities of different pre-trained layers. Moreover, the multi-layer protocol was also used by a lot of previous works \cite{DBLP:conf/eccv/NorooziF16,DBLP:conf/eccv/CaronBJD18,DBLP:conf/iclr/GidarisSK18}. We use exactly the same training configurations as in \footnote{https://github.com/open-mmlab/OpenSelfSup/blob/master/configs/\\benchmarks/linear\_classification/imagenet/r50\_multihead.py},  and no hyper-parameter tuning is performed. The initial learning rate is set to 0.1 with momentum 0.9 and weight decay 1e-4. The total fine-tuning epoch is set to 90, and the learning rate is decayed by 10 at epochs 30 and 60.

\textbf{Pascal VOC Object Detection:} We use the exact configurations as in \footnote{https://github.com/open-mmlab/OpenSelfSup/blob/master/bench-\\marks/detection/configs/pascal\_voc\_R\_50\_C4\_24k\_moco.yaml}. The Faster RCNN with ResNet-50-R4 backbone is trained on \texttt{trainval07+12} set and evaluated on \texttt{test2007} set. We train for 24K iterations using SGD optimizer with batch size 16 (2 per GPU). We use the base learning rate 0.02, perform warmup for 100 iterations, and divided it by 10 at iterations 18K and 22K.

\textbf{COCO Object Detection and Instance Segmentation:} We use the exact configurations as in \footnote{https://github.com/open-mmlab/OpenSelfSup/blob/master/bench-\\marks/detection/configs/coco\_R\_50\_C4\_2x\_moco.yaml}, which is the standard 2x schedule in \cite{wu2019detectron2}.

\section{Additional Experimental Results}

\textbf{The complete results of ImageNet Linear Classification:} Our main paper only reports the ImageNet linear classification results of conv4 and conv5 in Table 3. For completeness, we list the complete results from all residual layers in Table \ref{tab:full3} of this supplementary document.

\begin{table}
\begin{center}
\setlength\tabcolsep{4pt}
\begin{tabular}{l|ccccc}
\hline
\multirow{2}{*}{Method} & \multicolumn{5}{c}{ImageNet}\\

& conv1 & conv2 & conv3 & conv4 & conv5\\
\hline
Rand Init   & 11.4 & 16.2 & 13.5  & 9.1 & 6.5\\
Supervised  & 15.2 & 34.0 & 47.9 & 67.6 & 76.2\\
\hline
Relative-Pos \cite{DBLP:conf/iccv/DoerschGE15} & 14.8 & 31.3 & 45.8 & 49.3 & 40.2\\
Rotation-Pred \cite{DBLP:conf/iclr/GidarisSK18} & 12.9 & 34.3 & 44.9 & 55.0 & 49.1\\
\hline
NPID \cite{DBLP:conf/cvpr/WuXYL18} & 14.3 & 31.2 & 40.7 & 54.5 & 56.6\\
MoCo v2 \cite{DBLP:journals/corr/abs-2003-04297} & 14.7 & 32.8 & 45.0& \underline{61.6} & \underline{66.7}\\
SimCLR \cite{DBLP:conf/icml/ChenK0H20} & 17.1 & 31.4 & 41.4 & 54.4 & 61.6\\
BYOL \cite{DBLP:conf/nips/GrillSATRBDPGAP20} & 15.5 & 34.5 & 47.2 & \textbf{62.8} & \textbf{71.6}\\
\hline
MaskCo & 15.4 & 33.6 & 45.8 & 59.6 & 65.1\\

\hline
\end{tabular}
\end{center}
\caption{The complete ImageNet  linear classification results of Table 3 of the main paper.}
\label{tab:full3}
\end{table}

\begin{figure}[t]
\centering  
\includegraphics[width=0.9\linewidth]{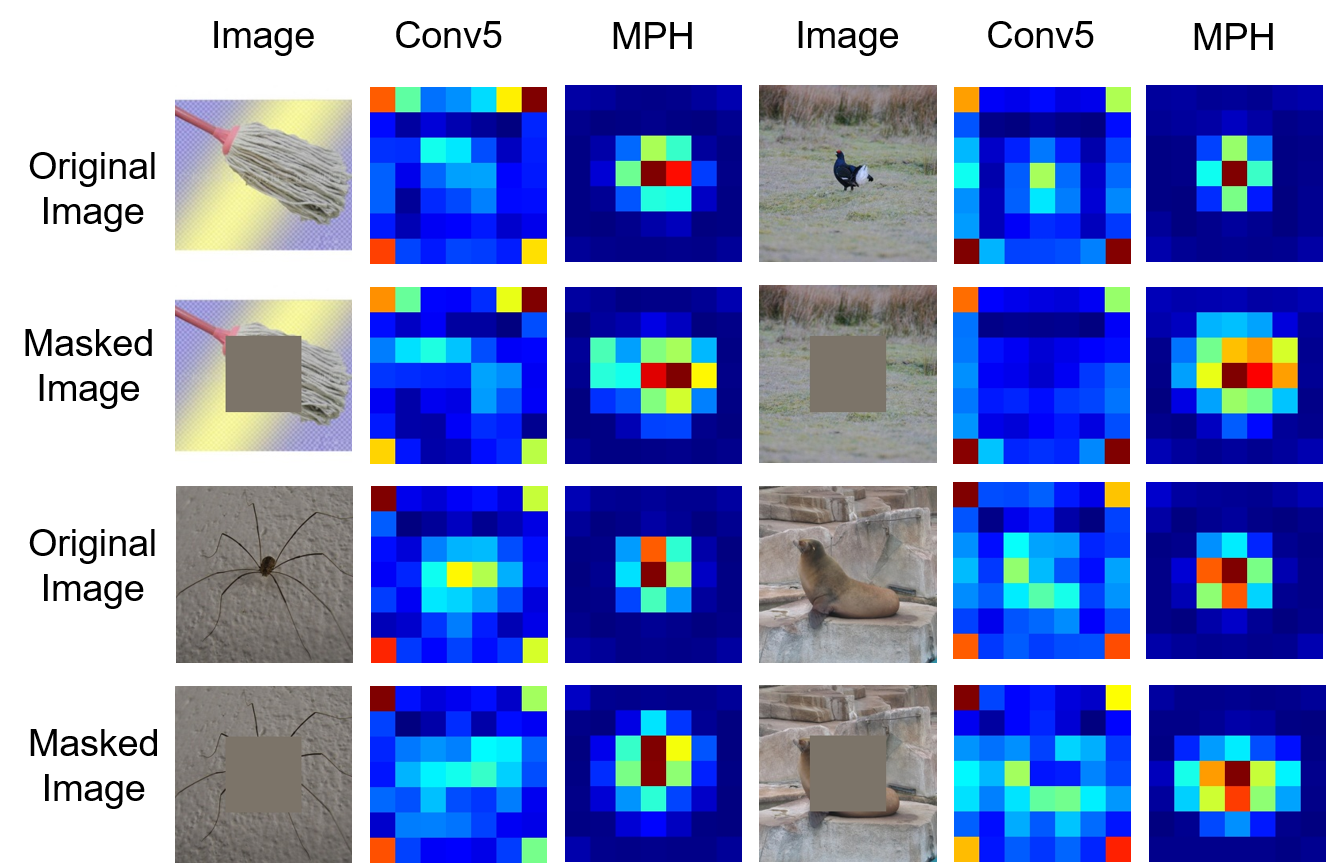}
\caption{Additional visualization results of MPH.}
\label{fig:morevis}
\end{figure}

\textbf{More Visualization results of MPH:} We also present additional visualization results of MPH in Figure \ref{fig:morevis}. 

\textbf{Ablation on COCO Pre-Training:} We report the ablation results of the mask strategy when our model is pre-trained on the COCO dataset in Table \ref{tab:coco4}. The trend is almost identical to what we find in ImageNet pre-training.

\begin{table}
\begin{center}
\begin{tabular}{c|ccc|cc}
\hline
\multirow{2}{*}{Mask}   & \multicolumn{3}{c|}{Pascal VOC} & \multicolumn{2}{c}{ImageNet}  \\
                        & $AP$ & $AP_{50}$ & $AP_{75}$    & conv4 & conv5 \\
\hline
\checkmark              & 56.2  & 81.4 & 62.6 & 55.3 & 38.5 \\
                        & 54.1  & 79.5 & 59.5 & 51.8 & 51.1 \\
\hline
\end{tabular}
\end{center}
\caption{Ablation studies on the mask strategy when our model is pre-trained on the COCO dataset. MPH is not used in these experiments.}
\label{tab:coco4}
\end{table}

{\small
\bibliographystyle{ieee_fullname}
\bibliography{egbib}
}

\end{document}